\documentclass[conference]{IEEEtran}
\IEEEoverridecommandlockouts
\usepackage{cite}
\usepackage{amsmath,amssymb,amsfonts}
\usepackage{algorithmic}
\usepackage{graphicx}
\usepackage{textcomp}
\usepackage{xcolor}
\usepackage{anyfontsize}
\usepackage{algorithm,algorithmic}
\def\BibTeX{{\rm B\kern-.05em{\sc i\kern-.025em b}\kern-.08em
    T\kern-.1667em\lower.7ex\hbox{E}\kern-.125emX}}

\begin{document}
\title{MecQaBot: A Modular Robot Sensing and Wireless Mechatronics Framework for Education and Research\\}

\author{\IEEEauthorblockN{Alice James\textsuperscript{1}\IEEEauthorrefmark{1}, Avishkar Seth\textsuperscript{1}, Subhas Mukhopadhyay\textsuperscript{1}}
\IEEEauthorblockA{\textit{Faculty of Science and Engineering\textsuperscript{1}} \\
\textit{Macquarie University}\\
Sydney, Australia \\
alice.james@mq.edu.au}

}

\maketitle

\begin{abstract}

We introduce MecQaBot, an open-source, affordable, and modular autonomous mobile robotics framework developed for education and research at Macquarie University, School of Engineering, since 2019. This platform aims to provide students and researchers with an accessible means for exploring autonomous robotics and fostering hands-on learning and innovation. Over the five years, the platform has engaged more than 240 undergraduate and postgraduate students across various engineering disciplines. The framework addresses the growing need for practical robotics training in response to the expanding robotics field and its increasing relevance in industry and academia. The platform facilitates teaching critical concepts in sensing, programming, hardware-software integration, and autonomy within real-world contexts, igniting student interest and engagement. We describe the design and evolution of the MecQaBot framework and the underlying principles of scalability and flexibility, which are keys to its success. Complete documentation: \textit{\textcolor{blue}{https://github.com/AliceJames-1/MecQaBot}}

\end{abstract}

\begin{IEEEkeywords}
\textit{mobile robotics, robot sensing applications, education robotics, raspberry pi, robot operating system}
\end{IEEEkeywords}

\section{Introduction}

The recent exponential growth in robotics and AI technology highlights the necessity of integrating evolving concepts into engineering curricula. The most effective way to learn is through hands-on experience, right? Engaging with these evolving concepts allows students to apply their knowledge in real-world scenarios and then return to theory for deeper understanding. In the ever-changing field of robotics, it is crucial to showcase its possibilities and foster independent exploration while offering essential mentorship and support to motivate students. However, many commercially available platforms are costly, lack customizability, and often fail to deliver the challenging, practical experiences and creativity students need. Additionally, online open-source materials can be disorganized and time-consuming, making integrating them into structured courses challenging for universities and organizations. 

\begin{figure}[htpb!]    
    \centering
    \includegraphics[width=1\linewidth]{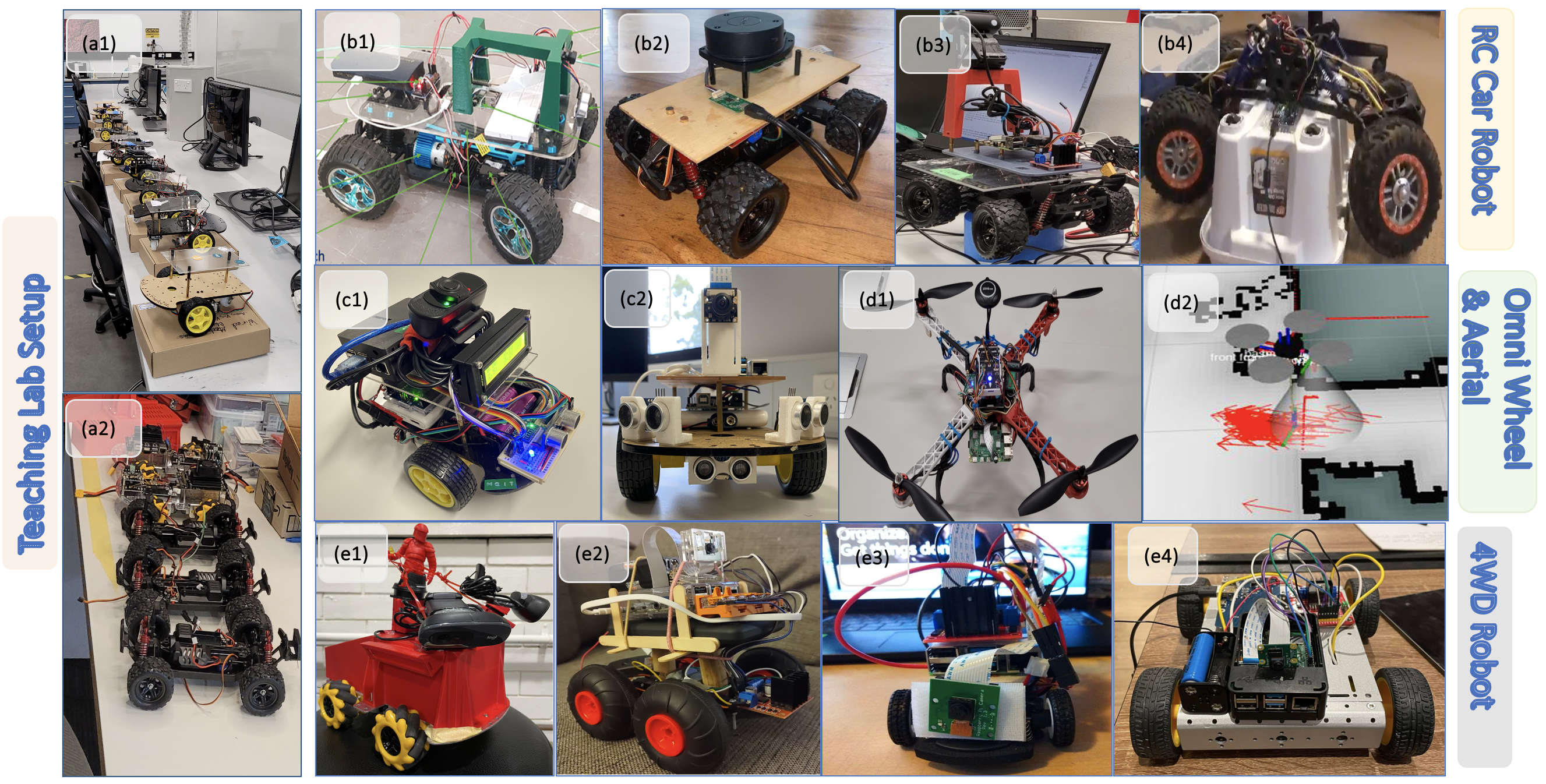}
    \caption{Photographs of the mobile robots built using the MecQaBots framework since 2020: 
    (a1) OmniWheel Chassis Teaching Lab Setup.
    (a2) RC Car Chassis Teaching Lab Setup.
    (b1) 1/10th Scale AGV (2019);
    (b2) LiDAR Mapping-Based AGV (2022); 
    (b3) RC Car CV AGV (2024); 
    (b4) Makeshift RC Car Arduino AGV (2021); 
    (c1) OmniWheel CV Bot (2023);
    (c2) Object Avoidance \& CV Bot (2023);
    (d1) Aerial Robot based on Raspberry Pi (2022);
    (d2) Gazebo Simulation of LiDAR-based Aerial Robot (2021);
    (e1) MURC Dual Camera Bot (2024);
    (e2) Truck Chassis CV Robot (2022);
    (e3) Raspberry Pi Tiny Robot Car (2021); 
    (e4) Differential 4WD Mobile Robot(2021); }
    \label{mecqabots}  
\end{figure}

We present MecQaBot, an open-source, cost-effective, and modular methodology for teaching autonomous mobile robotics developed by our research team at Macquarie University, Sydney, Australia. The MecQaBot is designed to be customizable and has several possibilities in robot types, as shown in Fig.~\ref{mecqabots}. The framework is a work in progress developed initially as a thesis project in 2019~\cite{Seth2022Ground, Seth2020Aug}. Since then, it has been used to teach over 240 undergraduate and postgraduate engineering students, mainly for the final year `Wireless Mechatronics' unit. The platform encourages learning state-of-the-art concepts in robotics, such as sensor integration, localization, navigation, computer vision and navigation. 

The MecQaBot framework utilizes open-source, cost-effective components for efficient prototyping. At its core is the Raspberry Pi computing board~\cite{James2021SepRPiIoT}, which offers a versatile range of configurations, from budget-friendly solutions with basic autonomous capabilities to advanced setups featuring 3D LiDAR or stereo cameras. The system operates alongside a wirelessly connected laptop or ground control device, which can run the ROS framework or utilize traditional Python/C++ APIs with or without Flask. The modular chassis design makes it easy to scale and modify hardware.

The contributions detailed in this work are as follows:
\begin{enumerate}
    \item A comprehensive collection of openly available resources designed for hobbyists, researchers, and educators seeking to explore robotics concepts.
    \item An in-depth overview of the hardware-software techniques and code utilized in mobile robotics, applicable across various scenarios.
    \item Multimedia and firsthand accounts from educators and students sharing their experiences and challenges encountered while prototyping these systems from the ground up.
\end{enumerate}

\section{Related Work}

Mobile robotics distinguishes itself from broader robotics disciplines by focusing on locomotive mechanisms enabling robots to move autonomously across different environments to fulfil diverse objectives. The concept of intelligent, autonomous mobile robots emerged 100 years ago, with significant advancements during the Second World War when they were crucial in military teleoperations. Since those early days, the integration of artificial intelligence (AI) and advances in sensor technology have revolutionized mobile robotics, expanding their applications across various fields, such as defence, logistics, healthcare, infrastructure inspection, swarming, and passenger transport. Today, mobile robots are increasingly autonomous, navigating complex environments with capabilities like path planning, obstacle avoidance, and localization~\cite{dudek2024computational, Moysiadis2020}. 

Mobile robots are crucial across various domains, classified by their operational environments: ground robots that use wheels for efficient movement, tracks for navigating rough terrain, and legs to mimic animal locomotion; aerial robots that exploit flight capabilities for defence, surveillance, entertainment, and delivery; aquatic robots designed for underwater or surface exploration and maintenance; and space and polar robots engineered for extreme extraterrestrial and polar conditions. These robots range from fully autonomous systems that utilize sensors and algorithms to perceive and interact with their environments to semi-autonomous variants that require human input for complex tasks~\cite{app12146951}. 

The evolution of mobile robotics has accelerated due to advancements in autonomy. Initially, robots were primarily teleoperated or followed simple pre-programmed instructions. This evolution began with Ray Goertz's development of the first teleoperated manipulator in 1948 for handling radioactive materials. Subsequent milestones, such as Cincinnati Milacron's mini-computer-controlled robot in 1971 and Stanford Research Institute's intelligent robot SHAKEY in 1972, laid the groundwork for autonomous functions in robotics~\cite{robotics9040109}. Significant developments include the Stanford Cart~\cite{stanfordcart}, Sojourner Rover~\cite{Wilcox1998JulSojounerMars}, BigDog Robot~\cite{Raibert2008JanBigdog}, Curiosity Mars Rover~\cite{hassler2014marscuriosity}, and Spot by Boston Dynamics~\cite{SpotBostonDynamics}. Each application not only demonstrates the versatility of mobile robots but also showcases their role in pushing the boundaries of exploration and improving operational safety and efficiency in challenging environments \cite{conry2013bioswimmer, manley2010wave, sharma2023review, caesar2020nuscenes}. 

Several notable open-source, ROS-based robots are utilized in academia and industry. Among these, projects such as DonkeyCar~\cite{donkeycar}, TurtleBot~\cite{Amsters2019Augturtlebot}, AWS DeepRacer~\cite{awsdeepracer}, Leo Rover~\cite{leorover}, Rover Eclipse~\cite{rovereclipse}, JPL Rover~\cite{jplrover}, Husarion Rosbot 2~\cite{husarionrosbot2}, Clover Drone~\cite{cloverdrone}, UQ Magni~\cite{uqmagni}, Crazyflie~\cite{crazyfliebit}, and Dexi~\cite{dexi} represent commercially available open-source initiatives, while Mini Pupper~\cite{minipupper}, F1/10~\cite{OKelly2020Aprf1tenth}, Duckietown~\cite{duckietown}, and MuSHR~\cite{Srinivasa2019Aug}, and MBot Michigan~\cite{MBotMichiganICRA} are open-source projects developed by universities and institutions. Additionally, the DJI Tello~\cite{djitello} offers a non-ROS, commercially available open-source platform that supports programming. The FIRST Robotics Competition offers another excellent opportunity for students to engage with mobile robots, utilizing LEGO\textsuperscript{TM} components~\cite{FirstRoboticsFRC}.

We present an alternative approach to the research community by drawing inspiration from earlier and ongoing work while emphasizing greater flexibility and modularity in our framework. We aim to ignite curiosity and promote self-directed learning among students. By providing a customizable platform, we empower learners to explore diverse robotic concepts and applications, fostering a hands-on, experiential learning environment. This initiative enhances technical skills and cultivates critical thinking and problem-solving abilities essential for future innovators in robotics.

\section{The Development of MecQaBot}

\begin{figure}[htpb!]    
    \centering
    \includegraphics[width=1\linewidth]{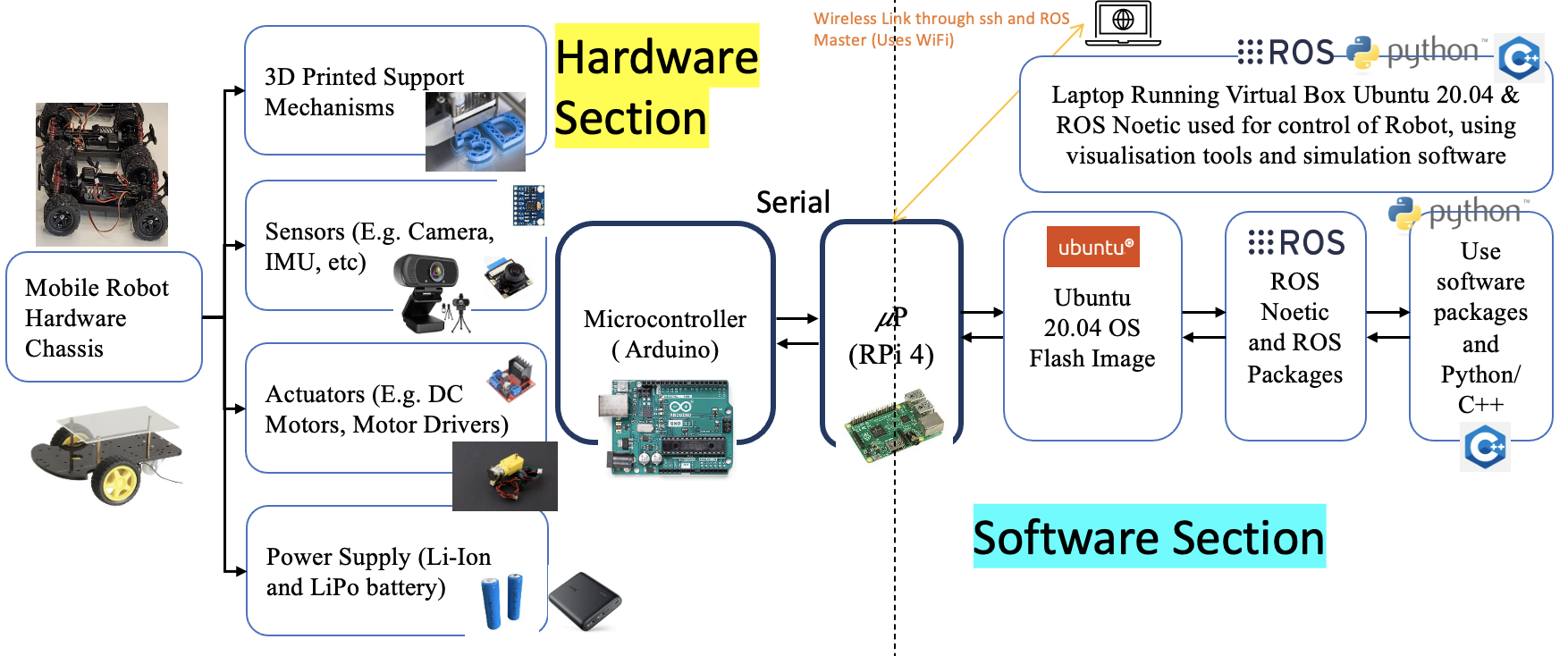}
    \caption{MecQaBot Hardware | Software System showing the connection between Raspberry Pi, Arduino, Sensors, Actuators, Wireless Network, and Software packages used.}
    \label{fig:rpi_sw-hw_sys}  
\end{figure}

In this section, we present an overview of the platform's hardware and software architectures. The hardware design is built from readily available off-the-shelf components, easily sourced online or in hardware stores worldwide. Meanwhile, the software architecture was developed at Macquarie University's research labs and is accessible through our GitHub page. Fig.~\ref{fig:rpi_sw-hw_sys} illustrates integrating hardware and software components in the MecQaBot system. 

\subsection{Hardware Architecture}

The hardware section includes a mobile robot chassis, either an OmniWheel or an RC car, enhanced with 3D-printed support mechanisms. Key components comprise sensors (e.g., a Logitech USB camera or Raspberry Pi camera), an RPLIDAR A1 LiDAR, and a BNO055 9DoF IMU. The chassis features generic DC motors included in the omnidirectional wheel robot kit, while the RC car typically has one DC motor for throttle and a servo motor for steering~\cite{Seth2020Aug}. The RC car may include an electronic speed controller (ESC) for throttle control. An L298N motor driver connects to the DC motors, equipped with encoders for the omnidirectional robot. The power supply consists of LiPo, NiMH or Li-Ion batteries, with a buck converter for continuous power to the components. The Raspberry Pi can also be powered separately using a mobile phone power bank.

These components can also connect to a microcontroller (Arduino), which communicates serially with a Raspberry Pi 4B. The Raspberry Pi typically runs Ubuntu 20.04 with ROS Noetic, interfacing wirelessly with a laptop running a virtual machine configured with the same software. Wireless communication between devices can be established using various technologies, including WiFi, Bluetooth, and LoRa, among many other options~\cite{James2021SepDesignIoT}. The Raspberry Pi 4B easily accommodates various operating systems (Ubuntu 20.04/ 22.04, Custom SD Card Image, or Raspberry Pi OS Bullseye) and programs via an SD card. This setup enables effective robot control, visualization, and simulation using software packages and programming languages like Python and C++.

\subsection{Software Architecture}

We use the Ubuntu 20.04 Focal Fossa Linux version in the MecQaBot robots. Linux can be used on Mac OS and Windows in different ways. A few options for using Linux Ubuntu on your Mac OS or Windows computer are available. One alternative is to use virtual machine software like VMware~\cite{VMWare2024} or Parallels~\cite{Parallels2024} for Mac or Oracle VM VirtualBox~\cite{VirtualBox2024} for Windows. Another option is to set up a dual boot system~\cite{DualBoot2024}, allowing you to run Ubuntu alongside your existing operating system. Windows users can also use WSL (Windows Subsystem for Linux)~\cite{WSL2024} for a more integrated Linux experience.
Additionally, Docker~\cite{Docker2024} provides containerized environments for both platforms. The specific method depends on the use case and application an individual wants to use it for. The dual boot Ubuntu Linux OS alongside a Windows OS is the best approach, providing users with ease of use and compatibility. Developing robotics applications requires integrating hardware peripherals, installing CPU-hungry software, and utilizing a good GUI that is easy to use with dual boot. 

\begin{figure}
    \centering
    \includegraphics[width=0.37\linewidth]{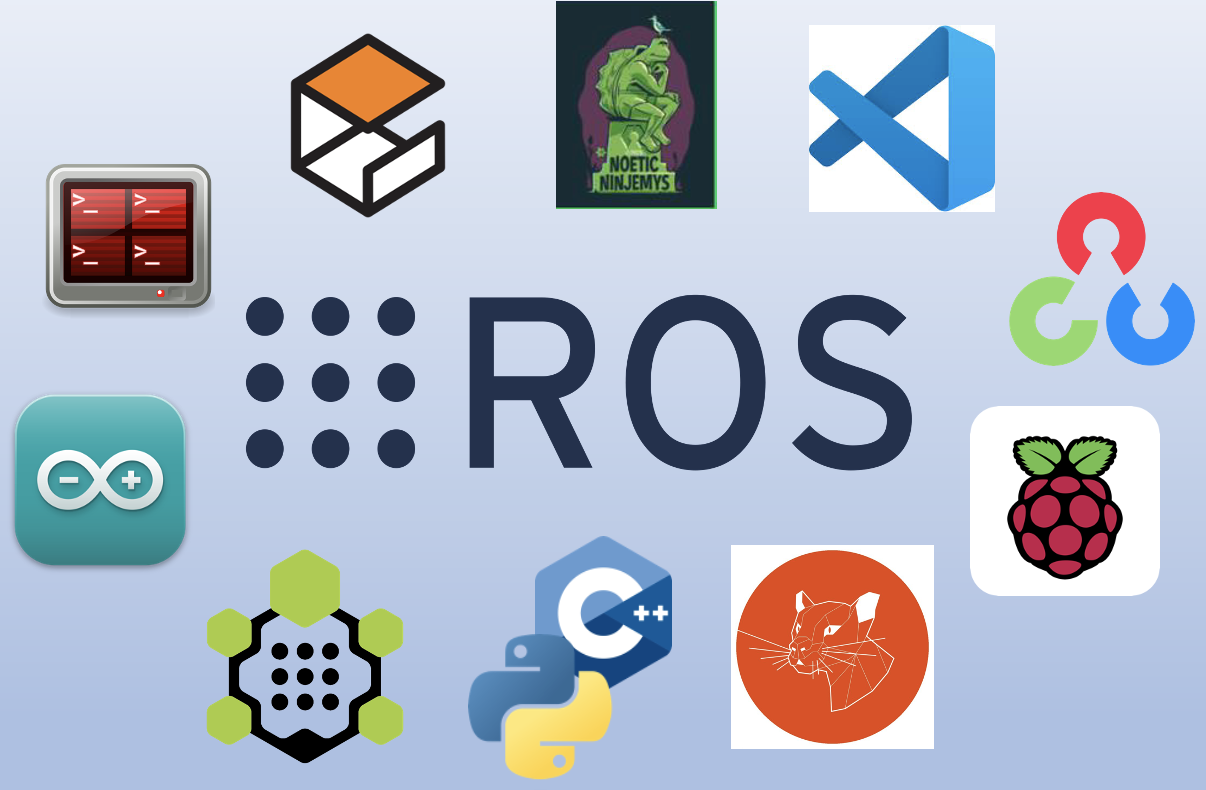}
    \includegraphics[width=0.61\linewidth]{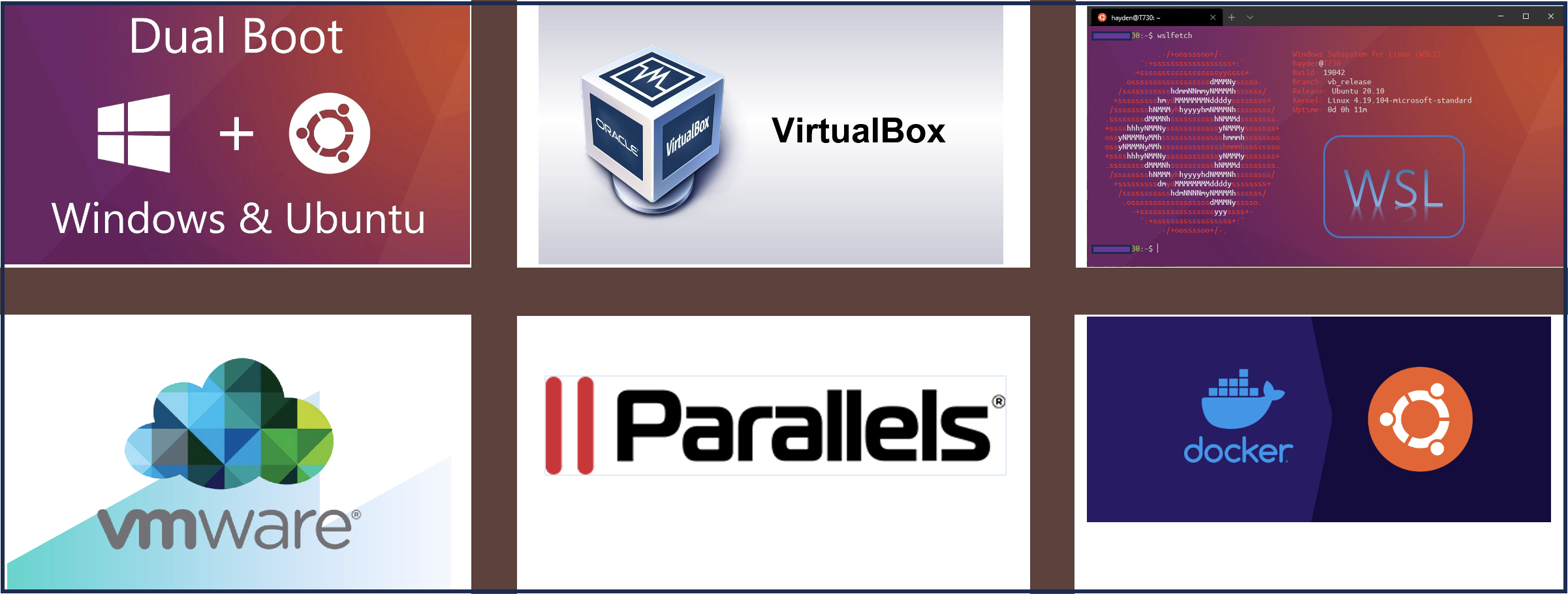}
    \caption{(a) ROS and its ecosystem with some of the major hardware-software components  used in our work (b) Different software applications that support using Linux OS on Windows or Mac OS}
    \label{fig:eco}
\end{figure}

The Robot Operating System (ROS) is a flexible framework for writing robot software. It is a collection of tools, libraries, and conventions that aim to simplify the creation of complex and robust robot behaviour across various robotic platforms~\cite{ROSWiki}. Figure~\ref{fig:eco} shows the ROS and its ecosystem with some of the significant hardware-software components used in our work. Some common elements used in the ROS system are microprocessors such as Raspberry Pi, microcontrollers such as Arduino, Gazebo simulator, and Python programming language.

\textbf{Robot Sensing}: Advanced autonomy relies heavily on multiple sensor inputs to implement SLAM (Simultaneous Localization and Mapping). Each sensor features an interface that converts raw inputs into ROS messages. Notably, the LIDAR and USB/RPi cameras have open-source ROS interfaces. Algorithms like gmapping, hector slam, and edge detection are used to navigate from point A to point B autonomously. Through serial communication or GPIO pins, PWM output values are sent to the actuators for navigation in real-world environments.

\section{Teaching with MecQaBot}

This section outlines the course structure, evaluates the teaching and student experiences, and discusses project demonstrations' technical challenges and outcomes. Recognizing that each learner has unique needs and varying learning speeds, the MecQaBot framework is designed to be both flexible and user-friendly. Its architecture allows users to choose interfaces tailored to their specific applications and skill levels, making the platform suitable for introductory and advanced users.

\subsection{Course Structure}

The lab sessions aim to provide students with hands-on experience in designing, building, and programming mobile robots using the Robot Operating System (ROS) and to develop skills in various areas, including computer vision, motor control, sensor integration, mapping, localization, and navigation. Ultimately, students will apply their knowledge to complete a mobile robot project and gain practical experience in robotics and autonomous systems.

\textbf{General Prerequisites}: Students should have a foundational understanding of the following areas:

\begin{itemize}
    \item Basics of Python/C++ and Linux (Ubuntu)
    \item Fundamentals of Electronics, Systems, Circuits, Sensors, and Motor Drivers
    \item Mathematics and Kinematics, including linear algebra and trigonometry
    \item An open mind and eagerness to learn
\end{itemize}

The course structure is devised with these prerequisites, building on mechanics, communications, software, and electronics concepts. 

\begin{figure}[htpb!]
\centering 
\includegraphics[width=1\linewidth]{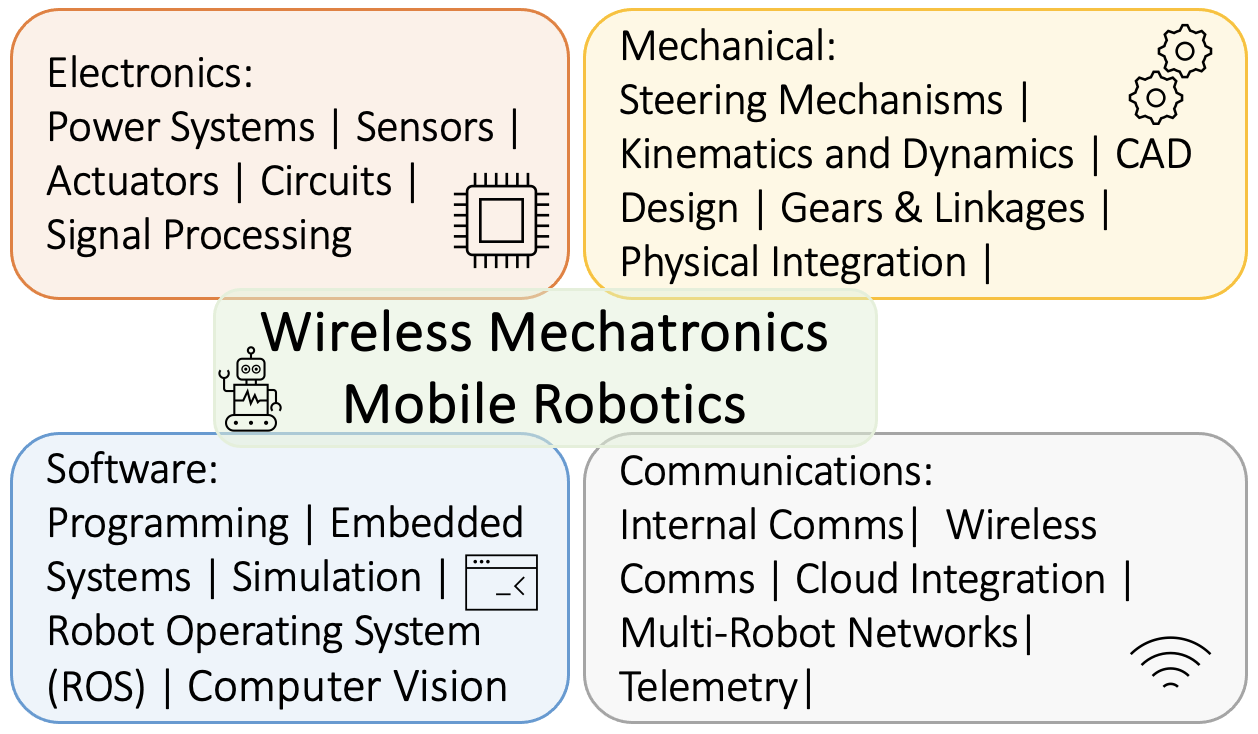} 
\caption{Core Components of Wireless Mechatronics | Mobile Robotics: A structured visualization of the essential areas—Electronics, Mechanical, Software, and Communications—illustrating the key concepts that integrate to form a complete robotic system.} 
\label{mtrn_rbtcs}
\end{figure}

Figure~\ref{mtrn_rbtcs} visually organizes the four core areas of Wireless Mechatronics and Mobile Robotics: Electronics, Mechanical Systems, Software, and Communications. Each section highlights fundamental concepts within these domains, such as sensors, kinematics, programming, and wireless protocols. This diagram serves as a concise roadmap for students to understand the multidisciplinary nature of mobile robotics.

Table~\ref {tab:schedule} outlines the course schedule for the MecQaBot program at Macquarie University, designed for undergraduate and postgraduate engineering students. This structured 13-week plan is a general method for teaching critical concepts in mobile robotics, emphasizing a hands-on approach to learning. Each week focuses on vital topics, beginning with foundational knowledge in robotics and Linux, followed by an in-depth exploration of the Robotics Operating System (ROS) and its applications. Students will engage with advanced concepts such as robot navigation and perception, integrating theoretical principles with practical skills. The latter part of the course is dedicated to project development, allowing students to apply their acquired knowledge in a collaborative environment. This framework not only fosters a comprehensive understanding of mobile robotics but also equips students with the necessary skills for real-world applications in engineering.

\begin{table}[ht]
  \caption{Course Schedule for Wireless Mechatronics Class}
  \label{tab:schedule}
  \resizebox{\columnwidth}{!}{
    \begin{tabular}{|l|c|} 
    \hline
    \textbf{Title} & \textbf{Keywords} \\ \hline
    Week 1: Introduction to Robotics & Overview, objectives, history, trends \\ \hline
    Week 2: Basics of Linux & Linux OS, commands, environment setup \\ \hline
    Week 3: Introduction to ROS & ROS architecture, installation, nodes, topics \\ \hline
    Week 4: ROS Tutorials Part I & Filesystem navigation, Python | C++ \\ \hline
    Week 5: ROS Tutorials Part II & Gazebo simulation, advanced concepts \\ \hline
    Week 6: Robot Navigation & SLAM, path planning, algorithms \\ \hline
    Week 7: Mobile Robot Design Principles & Design principles, control algorithms, integration \\ \hline
    Week 8: Sensors in Robotics & Camera setup, Lidar, sensor overview \\ \hline
    Week 9: Perception in Robotics & Computer vision, OpenCV, image processing \\ \hline
    Week 10: Teleoperation & Teleoperation setup, mobility testing \\ \hline
    Week 11: Project Development Week 1 & Hardware integration, basic programming \\ \hline
    Week 12: Project Development Week 2 & Algorithm refinement, perception features \\ \hline
    Week 13: Project Development Week 3 & Finalization, presentations, peer review \\ \hline
    \end{tabular}
  }
\end{table}

The table~\ref{tab:billofmaterials} presents the Bill of Materials for custom-built MecQaBot prototypes, highlighting three configurations: RC Car Basic Model, Omniwheel Basic Robot, and RC Car LiDAR, with all costs listed in AUD. Each configuration shares core components like the control board, battery, RPi 4 kit, USB camera, and additional sensors while differing in chassis type and wheels. LiDAR is included in the RC LiDAR version. The RC Basic and Omniwheel Basic configurations are more cost-effective, with total costs of 360 and 310, respectively. At the same time, the RC LiDAR version, equipped with a LiDAR sensor and a more advanced setup, reaches 500.

\begin{table}[ht]
  \caption{Bill of Materials for Custom Built MecQaBot Prototypes}
  \label{tab:billofmaterials}
  \resizebox{\columnwidth}{!}
  {
    \begin{tabular}{|l|c|c|c|c|} 
    \hline
    \textbf{Component}        & \textbf{Cost (AUD)} & \textbf{RC} & \textbf{Omniwheel} & \textbf{RC LiDAR} \\ \hline
    Control Board             & $30                  $ & $\checkmark$                            & $\checkmark$                               & $\checkmark$                            \\ \hline
    Motors                    & $10                  $ & |                            & $\checkmark$                               & |                            \\ \hline
    Battery                   & $30                  $ & $\checkmark$                            & $\checkmark$                               & $\checkmark$                            \\ \hline
    RC Car Chassis            & $120                 $ & $\checkmark$                            & |                               & $\checkmark$                            \\ \hline
    Omni wheel Chassis         & $40                  $ & |                            & $\checkmark$                               & |                            \\ \hline
    Basic Wheels              & $20                  $ & $\checkmark$                            & |                               & $\checkmark$                            \\ \hline
    Omni Wheels               & $10                  $ & |                            & $\checkmark$                               & |                            \\ \hline
    RPi 4 Kit                 & $100                 $ & $\checkmark$                            & $\checkmark$                               & $\checkmark$                            \\ \hline
    USB Camera                & $30                  $ & $\checkmark$                            & $\checkmark$                               & $\checkmark$                            \\ \hline
    LiDAR (optional)          & $200                 $ & |                            & |                               & $\checkmark$                            \\ \hline
    Additional Sensors        & $20                  $ & $\checkmark$                            & $\checkmark$                               & $\checkmark$                            \\ \hline
    \textbf{Total Cost}       &                      & \textbf{\$360}               & \textbf{\$310}                  & \textbf{\$500}               \\ \hline
    \end{tabular}
  }
\end{table}

\subsection{Teaching Evaluation}

\begin{figure}[htpb!]
\centering \includegraphics[width=0.43\linewidth]{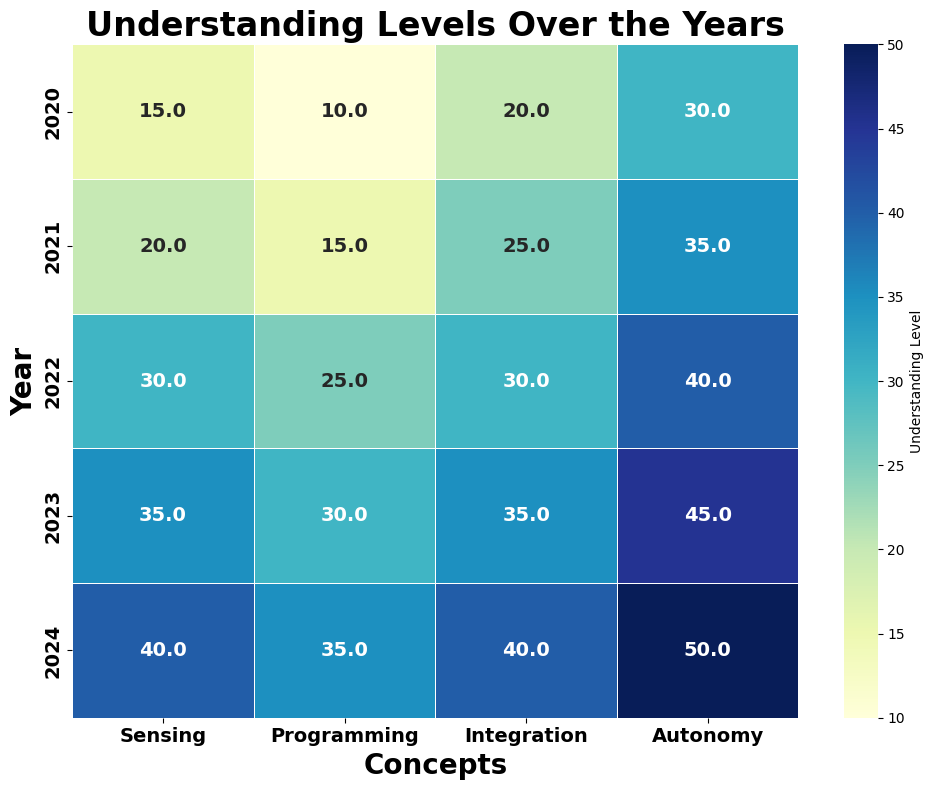} \includegraphics[width=0.55\linewidth]{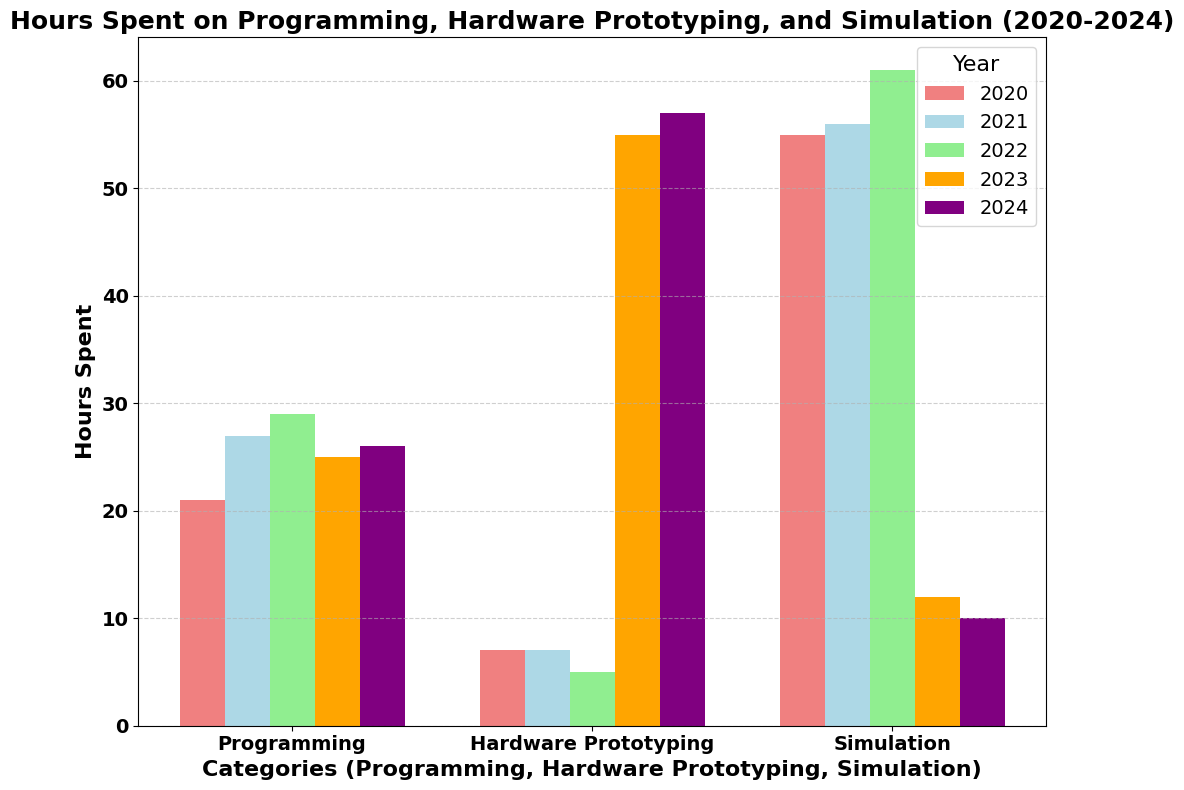} \caption{Heatmap and Grouped Bar Chart showing Understanding Levels and Time Allocation in WFH vs. Lab} 
\label{heatmap}
\end{figure}

The heatmap in Fig.~\ref{heatmap} represents students' understanding across four key concepts—Sensing, Programming, Integration, and Autonomy—over five years. The deeper shades signify higher levels of comprehension, illustrating a steady improvement in understanding. The data shows how students gradually adapted, particularly after the challenges posed by the COVID-19 pandemic in 2020-2021, which initially hindered progress. Annotations highlight key moments, such as the early difficulties and subsequent rise in mastery, culminating in significantly higher comprehension levels by 2024. The average understanding levels across each year further emphasize the collective growth in student knowledge.

The grouped bar chart in Fig.~\ref{heatmap} illustrates how students allocated their time across three primary topics— Programming, Hardware Prototyping, and Simulation—from 2020 to 2024. During the COVID-19 years (2020-2022), students spent significantly more time on simulation and programming while working from home, as lab access was restricted, leading to fewer hours on hardware prototyping. In the post-pandemic years (2023-2024), with a return to in-person learning, there was a marked increase in time spent on hardware prototyping in the lab, while simulation hours decreased. This shift highlights the impact of external factors like the pandemic on how students balanced theoretical and hands-on work over the years.

\subsection{Working Demonstration}

After nine weeks of learning and hands-on work in the lab, student teams are now tasked with presenting their functional robot models, each designed to perform various tasks for assessment. 
\begin{figure}[htpb!]
\includegraphics[width=0.95\linewidth]{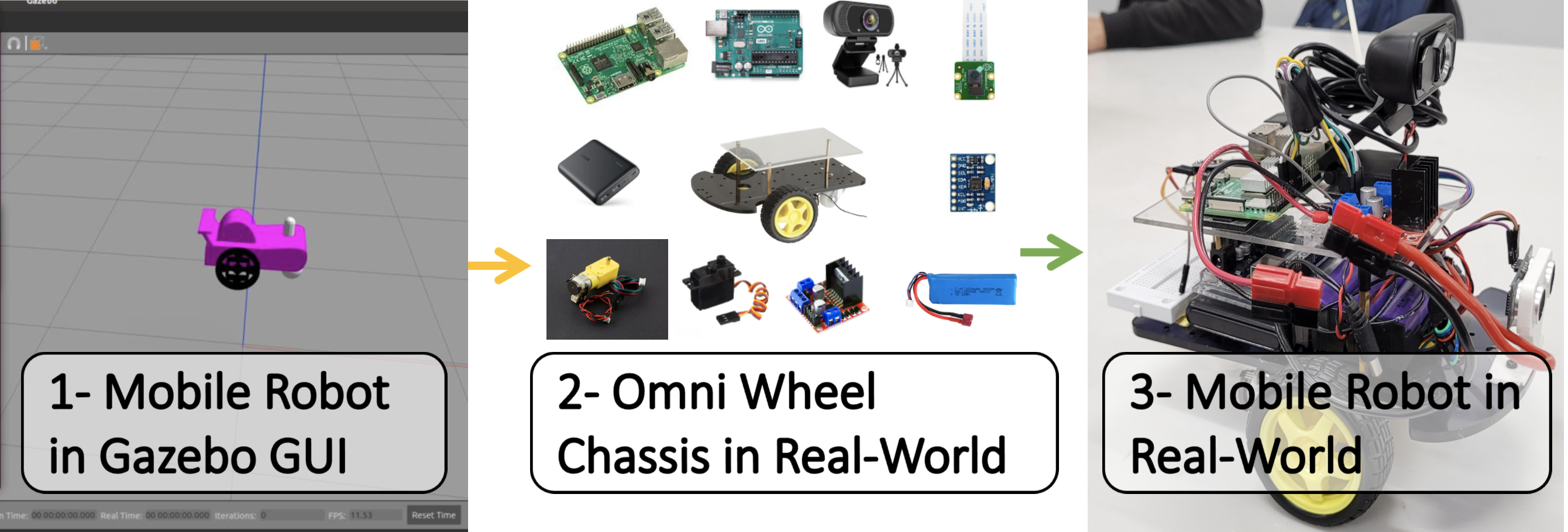} 
\centering \includegraphics[width=0.95\linewidth]{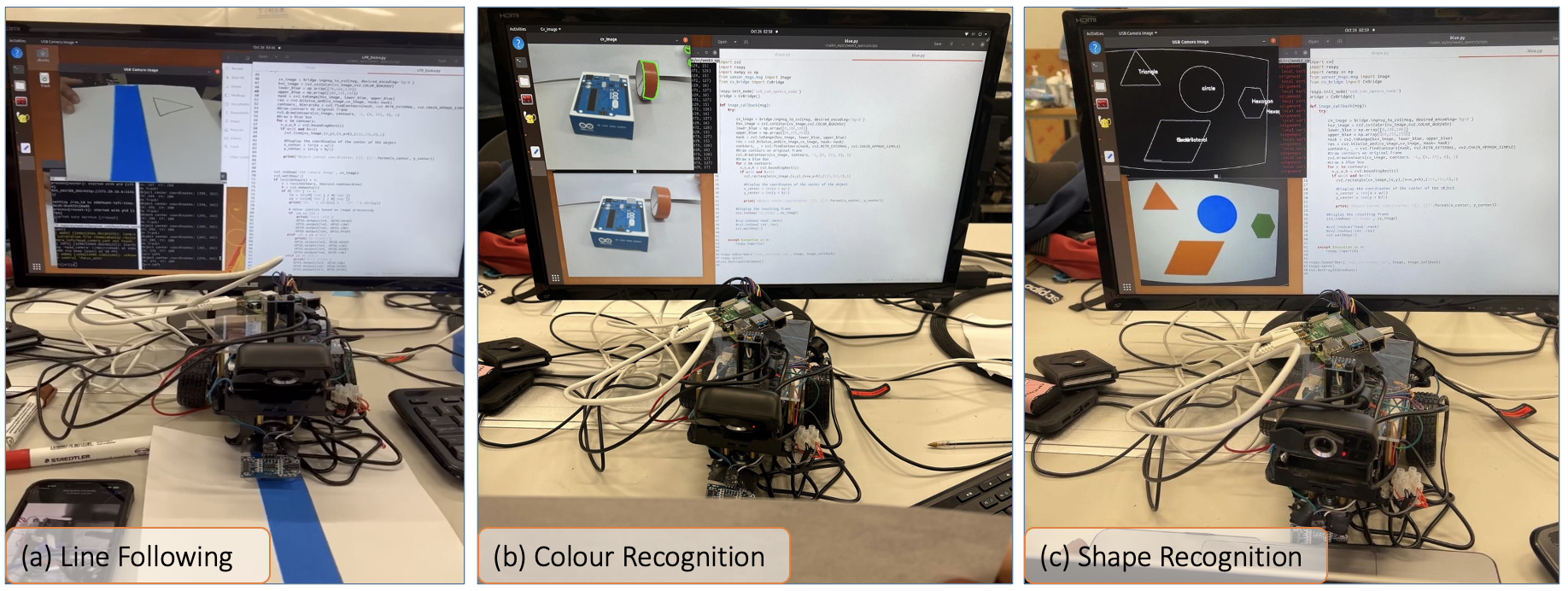} \includegraphics[width=0.95\linewidth]{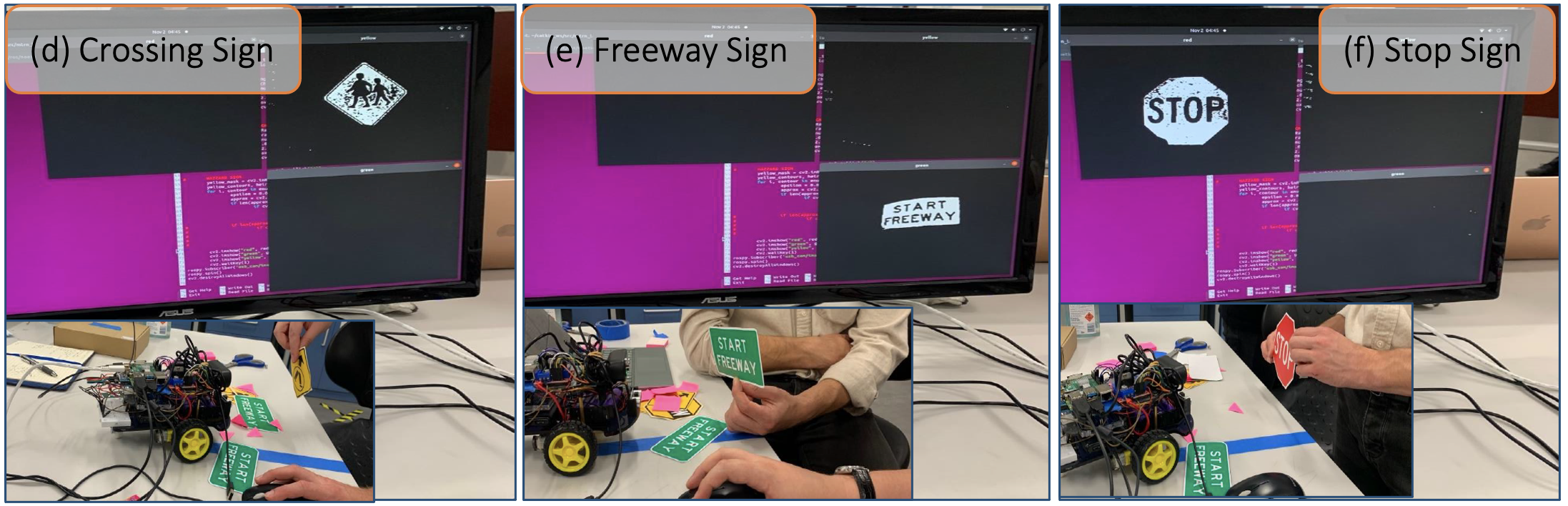} 
\centering \includegraphics[width=0.95\linewidth]{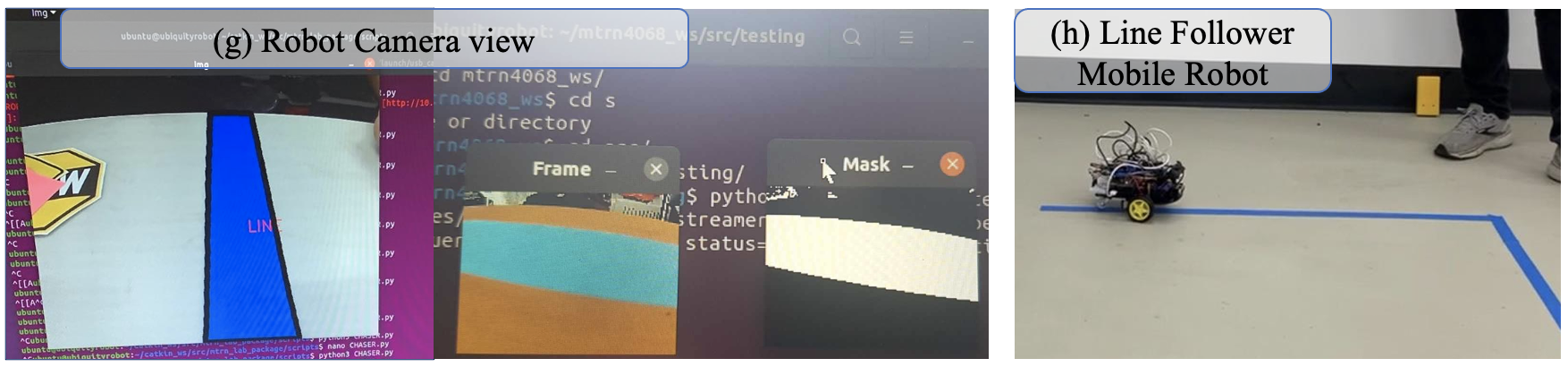} \includegraphics[width=0.95\linewidth]{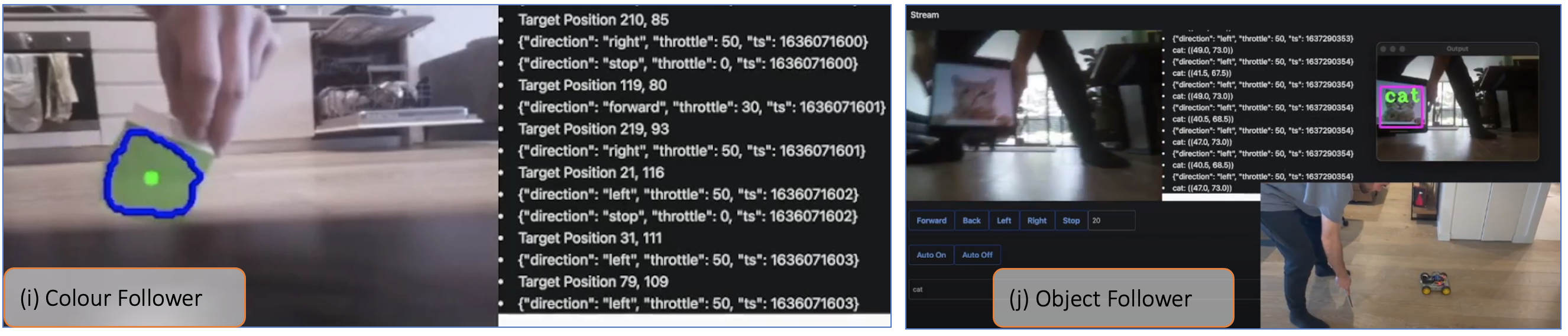} 
\includegraphics[width=0.95\linewidth]{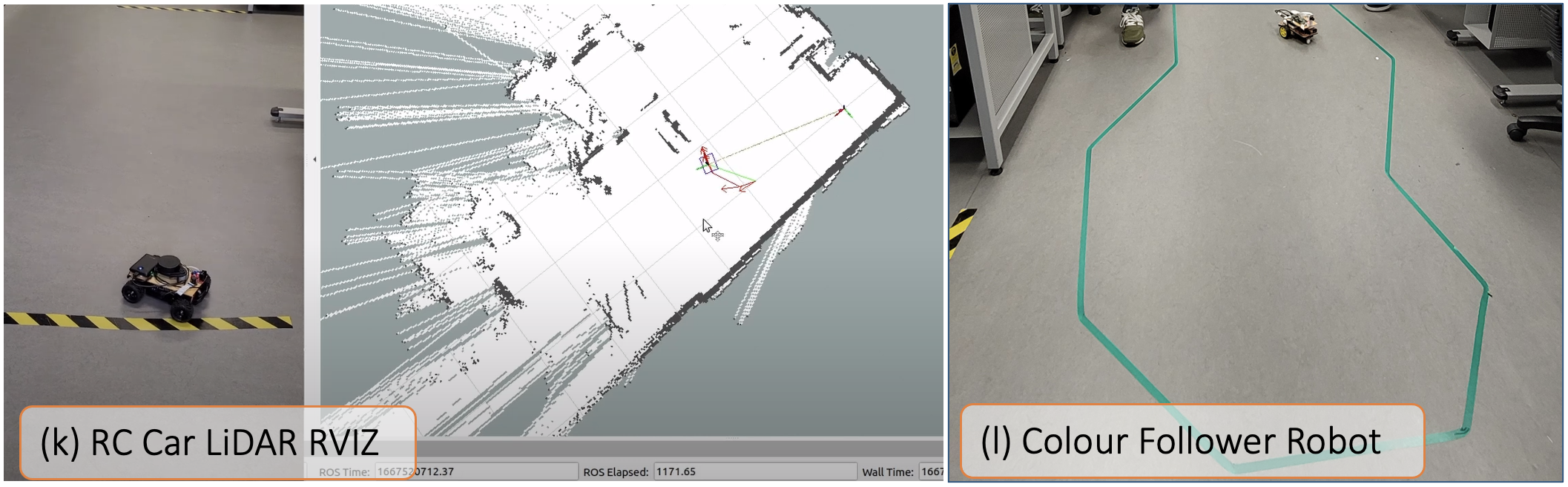}
\caption{Demonstration of Various Tasks in the Mechatronics Teaching Labs: Line Following (a, h), Color Recognition (b, i, l), Shape Recognition (c), Road Sign Detection (d, e, f), Robot Camera View (g), Object Following (j), and LiDAR Mapping with RViz (k) mapping its surroundings in 3D using RViz for spatial awareness and mapping for navigation (l) illustrates a color-following robot that tracks a green path.} 
\label{results}
\end{figure}
Figure~\ref{results} presents a comprehensive demonstration of custom-built mobile robots performing various tasks utilizing camera-based and LiDAR-based sensing capabilities. The robots engage in fundamental tasks such as line following, as shown in images (a) and (h), where the robot tracks a blue line using a vision-based algorithm. Images (b) and (i) display the robot's ability to recognize following colours, with additional examples of shape recognition in (c). In images (d), (e), and (f), the robot detects and responds to different road signs, such as a `Crossing' sign, a `Freeway' sign, and a `Stop' sign, indicating progress in real-time sign recognition. Image (g) provides a robot camera view, showcasing how the robot processes visual data. The object-following feature is highlighted in the image (j), where the robot detects and follows a moving object. A LiDAR-enabled RC car is presented in image (k), mapping its surroundings in 3D using RViz for spatial awareness and mapping for navigation. Lastly, image (l) illustrates a color-following robot that tracks a green path. These images demonstrate the robot's diverse perception, navigation, and environmental interaction capabilities.

\subsection{Overall Concluding Observations:}
\begin{itemize}
    \item \textbf{Technical Challenges:} 
    Hardware issues, software troubleshooting, wiring mistakes, and the iterative nature of the project highlighted the need for strong prerequisites and perseverance.
    
    \item \textbf{Student Experiences:}
    Students demonstrated creativity and self-directed learning while personalizing projects. Many found the tasks enjoyable despite challenges, acquiring transferable skills. Persistence often proved more valuable than technical knowledge, with some less experienced students outperforming their peers. 

    \item \textbf{Team Dynamics: }
    Working in teams of 2-3 presented task division and communication challenges. Effective delegation was crucial, with typical roles as follows: software development (40\%), mechanical design (30\%), and electronics/ comms design (30\%).

    \item \textbf{Project Management:} 
    Time commitment was essential; students learned to prioritize tasks and focus on outcomes over completion. This experience fostered critical thinking and reinforced that grades are not the sole measure of success in real-world projects.

    \item \textbf{Additional Insights:} 
    The importance of regular communication, conflict resolution strategies, leveraging individual strengths, constructive feedback, documentation, and adaptability to changing project requirements was learnt due to project-based learning.
\end{itemize}
\section{Conclusion}

We introduce MecQaBot, an innovative framework designed to enhance robotics education for both undergraduate and graduate students. Developed over nearly five years at Macquarie University, MecQaBot has proven effective across diverse student groups. This low-cost, open-source mobile robotic platform appeals to a broad audience, including hobbyists, educators, and researchers. Although it primarily focuses on autonomous driving, its capabilities can be adapted for other applications in fields such as aerial, marine, and space robotics. Additionally, we share insights into our teaching methods and student experiences, providing valuable information for future educators and learners. As the project evolves, we plan to collaborate with other institutions to integrate MecQaBot~\cite{MecQaBot} into more courses and expand its toolset, enhancing robotics education.

\section{Acknowledgements}
The authors express their gratitude to Macquarie University, Faculty of Science and Engineering, for providing lab spaces, equipment, and support personnel crucial to this project. We extend our heartfelt thanks to the participating students. Their contributions were invaluable, reminding us that teaching is a reciprocal process from which humans learn.

\bibliographystyle{IEEEtran}
\bibliography{refs_mtrn}

\end{document}